\documentclass[letterpaper, 10 pt, conference]{ieeeconf}  

\IEEEoverridecommandlockouts                              

\usepackage{graphics} 
\usepackage{subfig} 
\usepackage{wrapfig}

\usepackage{amsthm} 
\usepackage{amsmath,amssymb}

\usepackage{graphicx}
\usepackage{tabularx}
\usepackage{multirow, makecell}
\usepackage{diagbox}
\usepackage{slashbox}
\usepackage{rotating}
\usepackage{cite}
\usepackage{romannum}
\usepackage{url}
\usepackage{bm}
\usepackage[table]{xcolor}
\usepackage{hyperref}
\usepackage{siunitx} 
\usepackage{booktabs}
\usepackage{cleveref}
\usepackage{mathtools} 

\let\labelindent\relax 
\usepackage{enumitem} 

\usepackage[ruled,vlined]{algorithm2e}

\newcommand{\algovspace}{\vspace{1pt}}

\newcommand{\vx}{{\boldsymbol x}}
\newcommand{\vu}{{\boldsymbol u}}

\newcommand{\vz}{{\boldsymbol z}}

\newcommand{\StateSpace}{\mathcal{X}}
\newcommand{\ControlSpace}{\mathcal{U}}
\newcommand{\RealSpace}{\mathbb{R}}

\newcommand{\Rn}{\mathbb{R}^{n}}
\newcommand{\Rm}{\mathbb{R}^{m}}

\newcommand{\Natural}{\mathbb{Z}}
\newcommand{\calC}{\mathcal{C}} 
\newcommand{\calK}{\mathcal{K}} 

\newcommand{\calE}{\mathcal{E}} 
\newcommand{\calP}{\mathcal{P}} 
\newcommand{\calW}{\mathcal{W}} 

\newcommand{\calO}{\mathcal{O}} 
\newcommand{\calS}{\mathcal{S}} 
\newcommand{\calN}{\mathcal{N}} 
\newcommand{\calD}{\mathcal{D}} 

\newcommand{\vF}{{\bf F}}
\newcommand{\vE}{{\bf E}}
\newcommand{\vX}{{\bf X}} 
\newcommand{\vY}{{\bf Y}} 
\newcommand{\VSigma}{{\bm{\Sigma}}}
\newcommand{\vmu}{{\bm{\mu}}}

\newcommand{\horizon}{H} 

\DeclareMathOperator*{\argmin}{arg\,min}

\newtheorem{definition}{Definition}

\newtheorem{lemma}{Lemma}
\newtheorem{remark}{Remark}

\theoremstyle{definition}
\newtheorem{problem}{Problem}
\theoremstyle{definition}

\theoremstyle{definition}
\newtheorem{example}{Example}

\makeatother

\DeclareCaptionFont{mysize}{\fontsize{8}{9.6}\selectfont}
\title{\LARGE \bf Learning to Refine Input Constrained Control Barrier Functions via Uncertainty-Aware Online Parameter Adaptation}

\author{Taekyung Kim, Robin Inho Kee and Dimitra Panagou
\thanks{The authors would like to acknowledge the support of the National Science Foundation (NSF) through Award No. 1942907.}
\thanks{Taekyung Kim is with the Department of Robotics, University of Michigan, Ann Arbor, MI, 48109, USA {\tt\footnotesize taekyung@umich.edu}}
\thanks{Robin Inho Kee is with the Department of Mechanical Engineering, University of Michigan, Ann Arbor, MI, 48109, USA {\tt\footnotesize inhokee@umich.edu}}
\thanks{Dimitra Panagou is with the Department of Robotics and Department of Aerospace Engineering, University of Michigan, Ann Arbor, MI, 48109, USA {\tt\footnotesize dpanagou@umich.edu}}%
}

\begin{document}
\maketitle
\thispagestyle{empty}
\pagestyle{empty}

\begin{abstract}
Control Barrier Functions (CBFs) have become powerful tools for ensuring safety in nonlinear systems. However, finding valid CBFs that guarantee persistent safety and feasibility remains an open challenge, especially in systems with input constraints. Traditional approaches often rely on manually tuning the parameters of the class $\calK$ functions of the CBF conditions a priori. The performance of CBF-based controllers is highly sensitive to these fixed parameters, potentially leading to overly conservative behavior or safety violations. To overcome these issues, this paper introduces a learning-based optimal control framework for online adaptation of Input Constrained CBF (ICCBF) parameters in discrete-time nonlinear systems. Our method employs a probabilistic ensemble neural network to predict the performance and risk metrics, as defined in this work, for candidate parameters, accounting for both epistemic and aleatoric uncertainties. We propose a two-step verification process using Jensen-Rényi Divergence and distributionally-robust Conditional Value at Risk to identify valid parameters. This enables dynamic refinement of ICCBF parameters based on current state and nearby environments, optimizing performance while ensuring safety within the verified parameter set. Experimental results demonstrate that our method outperforms both fixed-parameter and existing adaptive methods in robot navigation scenarios across safety and performance metrics. \href{https://www.taekyung.me/online-adaptive-cbf}{\textcolor{red}{[Project Page]}}\footnote{Project page: \href{https://www.taekyung.me/online-adaptive-cbf}{https://www.taekyung.me/online-adaptive-cbf}} \href{https://github.com/tkkim-robot/online_adaptive_cbf}{\textcolor{red}{[Code]}} \href{https://youtu.be/255IUS1f6Lo}{\textcolor{red}{[Video]}}
\end{abstract}
\section{INTRODUCTION}

Safety-critical optimal control is one of the fundamental challenges in robotics, e.g., robots need to navigate safely around obstacles while optimizing various criteria such as energy consumption. Control Barrier Functions (CBFs) have been developed as a tool to encode safety for nonlinear systems and applied to high-level planning and multi-agent systems~\cite{ames_control_2014, ames_control_2017, ames_control_2019, li_survey_2023, garg_advances_2024, zhang_gcbf_2025, kim_visibility-aware_2025, glotfelter_nonsmooth_2021, lee_maintaining_2025}. High Order CBFs extend this concept to systems with constraints of higher relative degree with respect to the system dynamics~\cite{nguyen_exponential_2016, xiao_control_2019, breeden_high_2021, tan_high-order_2022}. However, a significant challenge lies in finding valid CBFs under input constraints~\cite{agrawal_safe_2021}. Functions that do not provide persistent safety and feasibility guarantees are often referred to as \textit{candidate} CBFs~\cite{tonkens_refining_2022}. %

Research in this field has taken three main directions. The first focuses on synthesizing CBFs directly~\cite{robey_learning_2020, tonkens_refining_2022, lee_data-driven_2023, ding_online_2024}, while the second learns CBFs through function approximation \cite{dawson_safe_2021, choi_robust_2021, liu_safe_2022, dawson_safe_2023, so_how_2024, lavanakul_safety_2024}. Another stream of research concentrates on tuning the class $\calK$ function parameters of CBFs. Many existing works verify and optimize CBF parameters offline and fix them during deployment~\cite{agrawal_safe_2021, xiao_feasibility-guided_2020, wang_ensuring_2022}. However, these fixed parameters may lead to overly conservative motions in new or changing environments. While several works introduce adaptive CBFs~\cite{taylor_episodic_2019, taylor_adaptive_2020, black_fixed-time_2021, xiao_adaptive_2022}, they address dynamics model uncertainty rather than safety and performance. %

Recent efforts in the third category have focused on adapting CBFs' class $\calK$ function parameters online \cite{parwana_trust-based_2022}. Some approaches~\cite{zeng_safetycritical_2021_decay, zeng_enhancing_2021} optimize decay rates of CBF constraints to guarantee point-wise feasibility, but fall short of guaranteeing safety. More recently, learning-based methods have become increasingly popular. However, their safety guarantees rely on unrealistic assumptions such as unbounded control inputs~\cite{ma_learning_2022} or perfect learned models~\cite{xiao_barriernet_2023, gao_online_2023}. %

To address these limitations, we propose a learning-based framework for online adaptation of CBF class $\calK$ function parameters in input-constrained systems. We leverage a probabilistic ensemble neural network~(PENN) to predict the safety and performance characteristics of potential parameter configurations while considering both epistemic and aleatoric uncertainties. The parameters are dynamically adapted based on the current system state and nearby environmental configurations, simultaneously ensuring safety and optimizing performance for navigation tasks. We quantify epistemic uncertainty, which originates from insufficient training data, using Jensen-Rényi Divergence, discarding predictions with low confidence. To account for aleatoric uncertainty, arising from sources such as data ambiguity and measurement noise, we model predicted outputs as distributions. We then employ distributionally robust Conditional Value at Risk to assess the risk associated with these predictions. %

We highlight the key contributions of this paper, which differentiate our approach from existing methods:
\begin{itemize}
\item We introduce a formal definition of discrete-time Input Constrained CBFs~(ICCBFs), specifically parameterized by class $\calK$ functions.
\item We propose a verification process for using a PENN model to predict the class $\calK$ functions of interest, ensuring \textit{robustness to both types of uncertainties}.
\item We develop an online adaptive MPC-ICCBF framework under \textit{input constraints}, dynamically adapting class $\calK$ functions to simultaneously ensure safety and optimize performance.
\end{itemize} %

\section{PRELIMINARIES \label{sec:preliminaries}}

Consider a general discrete-time nonlinear system:
\begin{equation}
\vx_{t+1} = f(\vx_t, \vu_t),
\label{eq:dynamics}
\end{equation}
where $\vx_t \in \StateSpace \subset \Rn$ is the state at time step $t \in \Natural^{+}$, and $\vu_t \in \ControlSpace \subset \Rm$ is the control input at time step $t$, with $\ControlSpace$ being a set of admissible controls for System~\eqref{eq:dynamics}. The function $f: \StateSpace \times \ControlSpace \to \Rn$ is assumed to be locally Lipschitz continuous with respect to both $\vx_t$ and $\vu_t$. 

\begin{definition}[Discrete-Time CBF \cite{agrawal_discrete_2017}]
Let $\calS = \{\vx \in \StateSpace \;|\; h(\vx) \geq 0\}$, where $h: \StateSpace \rightarrow \RealSpace$ is a continuous function. The function~$h$ is a discrete-time CBF for System~\eqref{eq:dynamics} if there exists a class $\calK$ function $\alpha(\cdot)$ satisfying $\alpha(z) < z$ for all $z > 0$ such that
\begin{equation}
\sup_{\vu_t \in \ControlSpace} \left[ \Delta h(\vx_t, \vu_t) \right] + \alpha(h(\vx_t)) \geq 0 \quad \forall \vx_t \in \StateSpace ,
\label{eq:cbf}
\end{equation}
where $\Delta h(\vx_t, \vu_t) = h(\vx_{t+1}) - h(\vx_t)$.
\label{def:cbf}
\end{definition}

\begin{lemma}
Given a discrete-time CBF~$h$ satisfying condition \eqref{eq:cbf} with the associated set~$\calS$, any control input $\vu_t \in K_\textup{cbf}(\vx_t)$, with $K_\textup{cbf}(\vx_t) \coloneqq \{\vu_t \in \ControlSpace : \Delta h(\vx_t, \vu_t) + \alpha(h(\vx_t)) \geq 0\}$, renders the set $\calS$ forward invariant for System~\eqref{eq:dynamics}. The proof can be found in \cite[Theorem 1]{ahmadi_safe_2019}.
\end{lemma} 

The discrete-time CBF condition ensures that the system state remains within the safe set $\calS$ for all future time steps, provided that the initial state lies within $\calS$.

\begin{remark}
The class $\calK$ function $\alpha$ can be chosen as a linear function with coefficient $\gamma \in (0, 1)$. In this case, it follows that $h(\vx_t) \geq (1 - \gamma)^t h(\vx_0)$ for all $t \in \Natural^{+}$. For $0 < \gamma \leq 1$, forward invariance of the set is still preserved (which is obviously true when $\gamma = 1$), and the discrete-time CBF becomes the discrete-time exponential CBF~\cite{agrawal_discrete_2017}.
\label{remark:linear_class_k}
\end{remark}

\section{PROBLEM FORMULATION \label{sec:problem}}
In this section, we formally define the discrete-time Input Constrained CBFs~(ICCBFs), and use this framework to formulate the main problems addressed in this paper.

\subsection{Input Constrained Control Barrier Functions \label{subsec:iccbf}}

The concept of CBFs has been generalized to HOCBFs in both continuous-time~\cite{xiao_control_2019} and discrete time~\cite{xiong_discrete-time_2023} settings, which can be used for constraints of high relative degree. ICCBFs~\cite{agrawal_safe_2021} further generalize HOCBFs, allowing for constraints of higher relative degrees and non-uniform relative degrees for control inputs.

More importantly, the ICCBF structure explicitly addresses the problem of generating CBFs under input constraints, i.e., $\ControlSpace \neq \Rm$. Concretely, suppose that the safe set~$\calS$ defined by $h$ cannot be rendered forward invariant by any feedback control input~$\vu \in K_\textup{cbf}(\vx)$, since there exist some states where it requires $\vu \notin \ControlSpace$ to render safe.

Using the same function $h$ in Definition~\ref{def:cbf}, we define a series of functions $b_i : \StateSpace \rightarrow \RealSpace$, $i = 0, \ldots, r$, as:
\begin{subequations}
\label{eq:iccbf-functions}
\begin{align}
b_0(\vx_t; \bm{\alpha}) &\coloneqq h(\vx_t) \\
b_1(\vx_t; \bm{\alpha}) &\coloneqq \inf_{\vu_t \in \ControlSpace} \left[ \Delta b_0(\vx_t, \vu_t; \bm{\alpha}) \right] + \alpha_0(b_0(\vx_t; \bm{\alpha})) \\
&\vdots \nonumber \\
b_r(\vx_t; \bm{\alpha}) &\coloneqq \inf_{\vu_t \in \ControlSpace} \left[ \Delta b_{r-1}(\vx_t, \vu_t; \bm{\alpha}) \right] + \alpha_{r-1}(b_{r-1}(\vx_t; \bm{\alpha})),
\end{align}
\end{subequations}
where $\bm{\alpha} = \{\alpha_0, \ldots, \alpha_r\}$ is a set of class $\calK$ functions with $\alpha_i : [0, a) \rightarrow [0, \infty)$ satisfying $\alpha_i(z) < z$ for $i = 0, \ldots, r$ and all $z > 0$, and $\Delta b_i(\vx_t, \vu_t; \bm{\alpha}) = b_i(\vx_{t+1}; \bm{\alpha}) - b_i(\vx_t; \bm{\alpha})$. We also define a series of sets $\calC_i$ associated with these functions as:
\begin{subequations}
\label{eq:iccbf-sets}
\begin{align}
\calC_0(\bm{\alpha}) &\coloneqq \{\vx_t \in \StateSpace : b_0(\vx_t; \bm{\alpha}) \geq 0\} = \calS \\
\calC_1(\bm{\alpha}) &\coloneqq \{\vx_t \in \StateSpace : b_1(\vx_t; \bm{\alpha}) \geq 0\} \\
&\vdots \nonumber \\
\calC_r(\bm{\alpha}) &\coloneqq \{\vx_t \in \StateSpace : b_{r}(\vx_t; \bm{\alpha}) \geq 0\} .
\end{align}
\end{subequations}

\begin{definition}[Inner Safe Set]
A non-empty closed set $\calC^*(\bm{\alpha}) \subseteq \calS$ is an inner safe set for System~\eqref{eq:dynamics} defined as~\cite{agrawal_safe_2021}:
\begin{equation}
\calC^{*}(\bm{\alpha}) \coloneqq \bigcap_{i=0}^{r} \calC_i(\bm{\alpha}) ,
\label{eq:inner-safe-set}
\end{equation}
and it is dependent on the class $\calK$ functions $\bm{\alpha}$.
\end{definition}

\begin{definition}[Discrete-Time ICCBF]
The function $b_r$ is a discrete-time ICCBF on $\calC^{*}(\bm{\alpha})$ for System~\eqref{eq:dynamics} if there exist class $\calK$ functions $\bm{\alpha} = \{\alpha_0, \ldots, \alpha_r\}$ such that
\begin{equation}
\sup_{\vu_t \in \ControlSpace} \left[ \Delta b_r(\vx_t, \vu_t; \bm{\alpha}) \right] + \alpha_{r}(b_r(\vx_t; \bm{\alpha})) \geq 0 \quad \forall \vx_t \in \calC^{*}(\bm{\alpha}) .
\label{eq:iccbf-condition}
\end{equation}
\label{def:iccbf}
\vspace{-12pt} 
\end{definition}

Note, the definition does not require condition~\eqref{eq:iccbf-condition} to hold for $\vx_t \in \calC_r(\bm{\alpha})$, instead, for a subset $\calC^{*}(\bm{\alpha}) \subset \calC_r(\bm{\alpha})$.

\begin{remark}
One can observe that the validity of the ICCBF function $b_r$ in satisfying condition~\eqref{eq:iccbf-condition} depends on the design of class $\calK$ functions~$\bm{\alpha}$.
\end{remark}

For notational simplicity, we denote the ICCBF constraint function as:
\begin{equation}
\psi(\vx_t, \vu_t; \bm{\alpha}) \coloneqq \Delta b_r(\vx_t, \vu_t; \bm{\alpha}) + \alpha_{r}(b_r(\vx_t; \bm{\alpha})) . 
\label{eq:iccbf-constraint}
\end{equation}

\begin{lemma}
Given the input constrained discrete-time system~\eqref{eq:dynamics}, if $b_r$ is a discrete-time ICCBF, then any control input $\vu_t \in K_\textup{iccbf}(\vx_t; \bm{\alpha})$, with $K_\textup{iccbf}(\vx_t; \bm{\alpha}) \coloneqq \{\vu_t \in \ControlSpace : \psi(\vx_t, \vu_t; \bm{\alpha}) \geq 0 \}$, renders the set $\calC^{*}(\bm{\alpha}) \subseteq \calS$ forward invariant. The proof is similar to \cite[Theorem 1]{agrawal_safe_2021}. 
\label{lemma:iccbf}
\end{lemma}

\begin{remark} 
We can solve the following optimization problem:
\begin{equation}
\zeta^{*} = \underset{\vx_t \in \calC^*(\bm{\alpha})}{\operatorname{minimize}} \sup_{\vu_t \in \ControlSpace} \left[ \Delta b_r(\vx_t, \vu_t; \bm{\alpha}) \right] + \alpha_{r}(b_r(\vx_t; \bm{\alpha})) .
\label{eq:validate}
\end{equation}
By the definition, $b_r$ is an ICCBF if and only if the optimization problem has a feasible solution $\zeta^{*} \geq 0$. However, this can only be used to invalidate $b_r$ as a valid ICCBF. The search over integers $r$ and class $\calK$ functions~$\bm{\alpha}$ to find valid ICCBFs can be complicated, similar to the challenge of finding Lyapunov functions in general. 
\end{remark}
\subsection{Problem Setup \label{subsec:problem}}

In the original ICCBF approach, which is common in many CBF applications, the objective is to identify a fixed set of class $\calK$ functions~$\bm{\alpha}$ that make the function $b_r$ a valid ICCBF for all states in the inner safe set $\calC^*$. However, this approach presents two main challenges: a) it is difficult to find such a set, and b) it can be overly conservative, potentially restricting the system's performance. 

To address these limitations, we propose an online adaptive approach that seeks to find a set of class $\calK$ functions that make the ICCBF candidate function $b_r$ to be locally valid~(to be formally defined in this section) at each control iteration, based on the current state and the surrounding environment. This method eliminates the need to exhaustively search over all possible class $\calK$ functions $\bm{\alpha}$ and validate \eqref{eq:validate} with the corresponding set $\calC^*(\bm{\alpha})$. Moreover, it allows the ICCBF to be dynamically tailored to the specific conditions the system encounters in real time.

Consider a navigation problem for a nonlinear system in the form of \eqref{eq:dynamics} in a static environment~$\calW$ containing information about a set of obstacles~$\calO$. We define this set as: $\calW = \{\calO_{j}: (\vz_{j}, l_{\textup{obs},j})\}_{j=1}^N$, where $\vz_j \in \RealSpace^2$ is the spatial coordinate and $l_{\textup{obs},j} \in \RealSpace^+$ is the radius of the $j$-th obstacle. Let us denote the MPC controller associated with an ICCBF~$b_r$ that depends on class $\calK$ functions~$\bm{\alpha}$ as $\pi_{\bm{\alpha}}: \calC^* \rightarrow \ControlSpace$ and $\vu = \pi_{\bm{\alpha}}(\vx) \in K_{\textup{iccbf}}(\vx; \bm{\alpha})$.
\noindent\rule{\columnwidth}{0.4pt}
\textbf{MPC-ICCBF:}
\begin{subequations}
\label{eq:mpc-iccbf-formulation}
\begin{align}
J^{*}(\vx_t; \bm{\alpha}) = \min_{\vu_{t:t+\horizon-1|t}} F(\vx_{t+\horizon|t})+ \notag \\
\sum_{k=0}^{\horizon-1} L(& \vx_{t+\tau|t}, \vu_{t+\tau|t}) \label{eq:mpc-iccbf-cost}  \\
\text{s.t.} \ \
\vx_{t+\tau+1|t} = f(\vx_{t+\tau|t}, \vu_{t+\tau|t}), \ & k = 0,...,\horizon-1 \label{eq:mpc-iccbf-dynamics} \\
\vu_{t+\tau|t} \in \ControlSpace, \ & k = 0,...,\horizon-1 \label{eq:mpc-iccbf-state-input-constraint} \\
\psi(\vx_{t+\tau|t}, \vu_{t+\tau|t}; \bm{\alpha}) \geq 0, \ & k = 0,...,\horizon-1. \label{eq:mpc-iccbf-constraint}
\end{align}
\end{subequations}
\noindent\rule{\columnwidth}{0.4pt}

The notation $\vx_{t+\tau|t}$ represents the predicted state vector at time step $t+\tau$, based on the information available at time step $t$. This prediction is obtained by initializing the state $\vx_{t|t} = \vx_t$ and applying the control sequence~$\vu_{t:t+N-1|t}$ to the system dynamics~\eqref{eq:mpc-iccbf-dynamics}. We use linear quadratic cost function both for the stage cost~$L$ and the terminal cost~$F$ in \eqref{eq:mpc-iccbf-cost} to navigate towards the goal location. The parameter $\horizon$ represents the prediction horizon. More details can be referred to in \cite{borrelli_predictive_2017, zeng_safetycritical_2021, thirugnanam_safety-critical_2022}.

Let us define the set of states of the system starting from a state $\vx_t$ for a fixed time $T_f$, driven by the controller $\pi_{\bm{\alpha}}$ parameterized by $\bm{\alpha}$:
\begin{equation}
\calP_t(\bm{\alpha}; \vx_t, \calE_t, \pi_{\bm{\alpha}}) \coloneqq \{\vx_{t+\tau}\}_{\tau=0}^{T_f}
\label{eq:trajectory_set}
\end{equation}
where $\vx_{t+\tau+1} = f(\vx_{t+\tau}, \pi_{\bm{\alpha}}(\vx_{t+\tau}))$. $\calE_{t} \subset \calW$ is the local map at time step $t$, defined as $\calE_t = \{\calO_j \in \calW : {\lVert \vz_t - \vz_j \rVert}_2 \leq l_{\textup{range}} + l_{\textup{obs},j}, j \in \{1, \ldots, N \}\}$, where $\vz_t \in \RealSpace^2$ is the robot's spatial coordinate, and $l_{\textup{range}}$ is a predefined sensing range.

\begin{definition}[Locally Valid Class $\calK$ Functions]
Given a set $\calP_t(\bm{\alpha}) \subset \calC^*(\bm{\alpha})$, the class $\calK$ functions $\bm{\alpha}$ are \textbf{locally valid} class $\calK$ functions for the candidate ICCBF $b_r$ on the set $\calP_t(\bm{\alpha})$ if the following condition is satisfied:
\begin{equation}
\zeta = \underset{\vx_t \in \calP_t(\bm{\alpha})}{\operatorname{minimize}} \sup_{\vu_t \in \ControlSpace} \left[ \Delta b_r(\vx_t, \vu_t; \bm{\alpha}) \right] + \alpha_{r}(b_r(\vx_t; \bm{\alpha})) \geq 0 .
\label{eq:local_validity}
\end{equation}
\end{definition}

Note, the local validity of the class $\calK$ functions~$\bm{\alpha}$ depends on current state~$\vx_t$, nearby environment~$\calE_t$, and the controller $\pi_{\bm{\alpha}}$ affected by $\bm{\alpha}$ itself.

\begin{example}
Consider a scalar discrete-time double integrator system with input constraints:
\begin{equation}
\begin{bmatrix} x_{t+1} \\
v_{t+1} \end{bmatrix} =
\begin{bmatrix} x_{t} + v_{t} \\
v_{t} + \vu_t \end{bmatrix}, \quad
\ControlSpace = [-1, 1].
\end{equation}
The safety constraint $\calS = \{\vx \in \RealSpace^{2} : x \geq -2\}$, i.e., $b_0(\vx_t; \bm{\alpha}) = h(\vx_t) = x_t + 2$, represents a wall-type obstacle located at $x = -2$. Now consider a candidate ICCBF $b_1(\vx_t; \bm{\alpha}) = \inf_{\vu_t \in \ControlSpace} \left[ \Delta b_0(\vx_t, \vu_t; \bm{\alpha}) \right] + \alpha_0(b_0(\vx_t; \bm{\alpha}))$ and linear class $\calK$ functions $\bm{\alpha} = \{ \gamma_0, \gamma_1 \}$ with $\gamma_0, \gamma_1 \in (0, 1)$. 

At $\calP_t = \{ \vx_t \}$ where $x_t = -1, v_t = -2$, the ICCBF constraint value in \eqref{eq:iccbf-constraint} with $\gamma_0 = \gamma_1 = 0.1$ is:
\begin{equation}
\sup_{\vu_t \in \ControlSpace} \left[ \Delta b_1(\vx_t, \vu_t; \bm{\alpha}) \right] + \gamma_{1} b_1(\vx_t; \bm{\alpha}) = \sup_{\vu_t \in \ControlSpace} \left[u_t \right] - 0.39 > 0,
\end{equation}
satisfying local validity. Contrary, $\gamma_0 = \gamma_1 = 0.5$ are not locally valid since $\sup_{\vu_t \in \ControlSpace} \left[u_t \right] - 1.75 < 0$. However, these class $\calK$ functions can be locally valid at a different state, e.g., $x_t = -1, v_t = -1$, as the value becomes $\sup_{\vu_t \in \ControlSpace} \left[u_t \right] - 0.75 > 0$.
\end{example}

\begin{problem}[Finding Locally Valid Class $\calK$ Functions]
Given the current state $\vx_t$, nearby environment $\calE_t$, and a pre-defined controller $\pi$, find a set of locally valid class $\calK$ functions for the candidate ICCBF $b_r$:
\begin{equation}
\mathfrak{A}_{\textup{valid}}(\vx_t, \calE_t, \pi) \coloneqq \{\bm{\alpha} : \zeta(\vx_t, \calE_t, \pi_{\bm{\alpha}}) \geq 0\}
\label{eq:valid_set}
\end{equation}
where $\zeta$ is defined as in \eqref{eq:local_validity}.
\label{prob:valid_cbf}
\end{problem}

Based on the solution of Problem~\ref{prob:valid_cbf}, we can ensure safety in the smaller inner set $\calP_t(\bm{\alpha}) \subset \calC^*(\bm{\alpha})$ by choosing locally valid class $\calK$ functions. As we iterate through the MPC process, we can update the set of locally valid class $\calK$ functions based on the current state and nearby environment. This allows us to focus on maximizing the MPC performance by selecting optimal parameters from the valid set.

\begin{problem}[Performance Optimization]
Given the set of locally valid class $\calK$ functions~$\mathfrak{A}_{\textup{valid}}(\vx_t, \calE_t, \pi)$, optimize the class $\calK$ functions such that maximizes the performance by minimizing the MPC objective function~\eqref{eq:mpc-iccbf-cost}. Formally, solve at each time step $t$:
\begin{equation}
\bm{\alpha}_t^{*} = \argmin_{\bm{\alpha} \in \mathfrak{A}_{\textup{valid}}(\vx_t, \calE_t, \pi)} J^*(\vx_t; \bm{\alpha})
\label{eq:performance_optimization}
\end{equation}
\label{prob:optimize_performance}
\end{problem}

\section{METHODOLOGY \label{sec:method}}
\subsection{Safety Loss Density Function \label{subsec:safety-loss}}

In this work, we assume a robot modeled as a dynamic unicycle with state $\vx_t = [x_t, y_t, \theta_t, v_t]^\top$, where $x_t$ and $y_t$ represent the position, $\theta_t$ is the heading angle, and $v_t$ is the linear velocity, and the control inputs $\vu_t = [a_t, \omega_t]^\top$ are the acceleration and the angular velocity, respectively.

In collision avoidance scenarios in robot navigation, the distances to the obstacles are not the sole factor of safety. The safety hard constraint for obstacle collision avoidance can be simply encoded as the signed distance function:
\begin{equation}
\label{eq:hard-constraint}
h(\vx_t) = (x_t-x_{\textup{obs}})^2 + (y_t-y_{\textup{obs}})^2 - (l_{\textup{robot}} + l_{\textup{obs}})^2 \geq 0 ,
\end{equation}
where \( x_{\textup{obs}} \) and \( y_{\textup{obs}} \) are the coordinates of the obstacle. The terms \( l_{\textup{robot}} \) and \( l_{\textup{obs}} \) represent the radii of the robot and the obstacle, respectively. It is obvious that a controller with a hard constraint like \eqref{eq:hard-constraint} cannot ensure safety from all initial conditions. The ICCBF constraint value in \eqref{eq:mpc-iccbf-constraint} is a useful metric to measure the safety margin with respect to the current robot state~$\vx_t$ and the system dynamics~\eqref{eq:dynamics}. However, the rate of change of the original hard constraint~$h$ in a discrete-time setting is computed via the time difference of subsequent value, it does not directly encode the direction of the robot's motion. Therefore, inspired by \cite{bena_safety-aware_2023}, we introduce a safety loss density function that incorporates both spatial state and directional information to provide a more comprehensive measure of collision risk.

Let $\Phi: \StateSpace \times \ControlSpace \to \RealSpace$ be a positive semi-definite density function that describes the safety loss of a particular robot state~$\vx$ and control input~$\vu$:
\begin{equation}
\Phi(\vx, \vu) \coloneqq \max_{j} \frac{\lambda_j(\vx, \vu)}{\beta_j(\vx) d(\vz, \vz_{j})^2 + 1} ,
\label{eq:safety_density}
\end{equation}
where $d(\vz, \vz_{j}) \coloneqq {\lVert \vz - \vz_j \rVert}_2 - l_{\textup{robot}} - l_{\textup{obs}}$, and $\vz, \vz_{j} \in \RealSpace^2$ are the spatial location of the robot and the $j$-th among $N$ obstacles. 

$\lambda_j: \StateSpace \times \ControlSpace \to \RealSpace$ determines the peak height of the density function, and it is defined as:
\begin{equation}
\lambda_j(\vx, \vu) \coloneqq \lambda_{1} e^{-\lambda_{2} \psi_j(\vx, \vu; \bm{\alpha}) },
\label{eq:alpha_j}
\end{equation}
where $\psi_j(\vx, \vu; \bm{\alpha})$ is the ICCBF constraint value~\eqref{eq:mpc-iccbf-constraint} for the $j$-th nearby obstacle. We use $r=1$ to construct the ICCBF candidate given that the function $h$ has relative degrees of $2$ for both control inputs. $\lambda_{1} \in \RealSpace$ determines the nominal peak height, and $\lambda_{2} \in \RealSpace$ defines the risk decay rate.

$\beta_j: \StateSpace \to \RealSpace$ influences the dissipation rate of the density function based on the directional motion of the robot:
\begin{equation}
\beta_j(\vx) \coloneqq \beta_{1} e^{\beta_{2} (\cos(\Delta\theta_j) + 1)},
\label{eq:beta_j}
\end{equation}
where $\Delta\theta_j$ is the angle between the robot's motion direction and the vector pointing from the robot to the $j$-th obstacle. $\beta_{1},\beta_{2} \in \RealSpace$ are the parameters similar to \eqref{eq:alpha_j}.

By defining \eqref{eq:safety_density}-\eqref{eq:beta_j}, we can measure the collision risk based on the robot state.
\subsection{Training Data Generation - Offline Sampling\label{subsec:data-gen}}

To train a prediction model that can infer the safety and performance of the MPC-ICCBF controller based on the state of the robot and the choice of the class $\calK$ functions, we collect a dataset~$\calD$ by generating robot trajectory $\calP_0(\bm{\alpha}; \vx_0, \calE_0, \pi_{\bm{\alpha}})$ as in \eqref{eq:trajectory_set} offline, similar to \cite{xiao_feasibility-guided_2020}.


We uniformly sample the initial state of the robot $\vx_0 = [x_0, y_0, \theta_0, v_0]^\top$ and the positions of $N$ obstacles, while at least one obstacle is placed in the path towards the fixed goal location, requiring the robot to actively avoid it. We also sample linear class $\calK$ functions $\gamma_0, \gamma_1$ according to Remark~\ref{remark:linear_class_k}. The initial distance to obstacle~$d_0$ is computed as $d(\vz_{0}, \vz_{j})$, where $\vz_{0} = [x_0, y_0]^\top$ and $\vz_{j}$ is the location of the obstacle that imposes the greatest safety loss value in \eqref{eq:safety_density} at $\vx_0$ and $\vu = [0, 0]^\top$. $\Delta\theta_0$ represents the relative angle as defined as in \eqref{eq:beta_j}.

Based on the configuration above $(d_0, \Delta \theta_0, v_0, \gamma_0, \gamma_1)$, we navigate the robot by solving the MPC-ICCBF problem~\eqref{eq:mpc-iccbf-formulation} and record two metrics as the ground truth for prediction. To solve Problem~\ref{prob:valid_cbf}, we compute the risk level~$\phi \in \RealSpace$: $\phi = \max_{t \in [0, T_{\textup{max}}]} \Phi(\vx_t, \vu_t)$, where $T_{\textup{max}}$ is the maximum simulation time. To approximately solve Problem~\ref{prob:optimize_performance}, we compute the deadlock time~$\delta \in \RealSpace$, which is an accumulated time segment where the robot exhibits a smaller velocity than a pre-defined threshold $v_{\textup{thr}}$. If the robot remains stationary until $T_{\textup{max}}$, we set it as the maximum value~$\delta = \delta_{\textup{max}}$. A training data pair consists of the input $X = [d_0, \Delta \theta_0, v_0, \gamma_0, \gamma_1]^\top$ and the ground truth $Y = [\phi, \delta]^\top$.

\subsection{Probabilistic Ensemble Neural Network Model \label{subsec:data-train}}

We use Probabilistic Ensemble Neural Network~(PENN), denoted as $\vF$, to predict the output vector~$Y = [\phi, \delta]^\top$. It offers a principled way to quantify both aleatoric and epistemic uncertainties~\cite{chua_deep_2018, buckman_sample-efficient_2018, kim_bridging_2023}.

Each neural network model~$\vE_{b}$ of the ensemble members outputs a Gaussian distribution conditioned on the model input~$X$:
\begin{equation}
    \vE_{b}(X; \bm{\theta}_{b}) \sim \calN(\vmu_{\bm{\theta}_{b}}(X), \VSigma_{\bm{\theta}_{b}}(X))  \,,
    \label{eq:gaussian}
\end{equation}
where $\vmu$ and $\VSigma$ are the mean and the diagonal covariance parameterized by the model parameter~$\bm{\theta}_{b}$. Given a collected dataset~$\calD$, each probabilistic model is trained with the Gaussian Negative Log-Likelihood~(NLL) loss function~\cite{kim_bridging_2023}. Given the ensemble of $B$ models~$\mathfrak{E} = \{\vE_{1}, \ldots, \vE_{B} \}$, $B\geq2$, the predicted vector~$\hat{Y}$ follows a Gaussian Mixture Model~(GMM) distribution:
\begin{equation}
    \hat{Y} \sim \vF(X ; \bm{\theta}_{1:B}) = \sum_{b=1}^{B}{w_b \vE_{b}(X; \bm{\theta}_{b})} \,, \, 0 \leq w_{b} \leq 1 .
    \label{eq:penn}
\end{equation}
Here, we simply use equal weights~$w_b = \frac{1}{B}, b = 1, \ldots, B$. 

Since the initial weights of each ensemble member are randomly initialized individually, disagreement among the predictions serves as a measure of epistemic uncertainty. 
\subsection{Uncertainty-Aware Class $\calK$ Functions Verification \label{subsec:assess}}

Now we propose the solution for Problem~\ref{prob:valid_cbf} by considering both types of uncertainties. 

\subsubsection{Epistemic Uncertainty Quantification}

We evaluate the \textit{confidence} of the prediction by measuring epistemic uncertainty. We employ Jensen-Rényi Divergence~(JRD) with quadratic Rényi entropy~\cite{renyi_measures_1961}, which has a closed-form expression of the divergence of a GMM~\cite{wang_closed-form_2009, kim_bridging_2023}:
\begin{align}
    D(X; \mathfrak{E}) & = - \log \left( \frac{1}{B^{2}}  \sum_{b,c}^{B} \mathfrak{D}_{(b,c)} \right) +  \frac{1}{B} \sum_{b}^{B} \log \left[ \mathfrak{D}_{(b,b)} \right] , 
    \label{eq:epistemic_uncertainty} \\
    \mathfrak{D}_{(b,c)} &\coloneqq \frac{1}{|\bm{\mathfrak{V}}|^{\frac{1}{2}}} \exp \left(-\frac{1}{2} \bm{\Delta}^{\top} \bm{\mathfrak{V}}^{-1} \bm{\Delta}\right),
\end{align}
$\bm{\mathfrak{V}} \coloneqq \VSigma_{\bm{\theta}_{b}}+\VSigma_{\bm{\theta}_{c}}$ and $\bm{\Delta} \coloneqq \vmu_{\bm{\theta}_{b}}-\vmu_{\bm{\theta}_{c}} $. If the JRD~$D(X)$ of the prediction of a given input~$X$ is greater than the pre-defined threshold~$D_{\textup{thr}}$, it is deemed to be out-of-distribution.

\subsubsection{Distributionally Robust Risk Assessment}

We then evaluate the risk of the predicted risk level~$\hat{\phi} \in \hat{Y}$, which follows a GMM distribution due to aleatoric uncertainty. To this end, we first establish the connection between the predicted risk level and the local validity of class $\calK$ functions. The condition~\eqref{eq:local_validity} can be extended as:
\begin{align}
\zeta & = \underset{\vx_t \in \calP_t(\bm{\alpha})}{\operatorname{minimize}} \sup_{\vu_t \in \ControlSpace} \left[ \Delta b_r(\vx_t, \vu_t; \bm{\alpha}) \right] + \alpha_{r}(b_r(\vx_t; \bm{\alpha})) \\
& \geq \underset{\vx_t \in \calP_t(\bm{\alpha})}{\min} \, \psi(\vx_t, \vu_t; \bm{\alpha}) \geq 0 .
\label{eq:condition1}
\end{align}
By further transforming this condition~\eqref{eq:condition1} and using the definition of safety loss density function~$\Phi$~\eqref{eq:safety_density} and risk level~$\phi = \max_{t \in [0, T_{\textup{max}}]} \Phi(\vx_t, \vu_t)$, we can derive:
\begin{align}
& \phi = \max_{\vx_t \in \calP_t(\bm{\alpha})} \max_{j} \frac{\lambda_1 e^{-\lambda_2 \psi(\vx_t, \vu_t; \bm{\alpha})}}{\beta_1 e^{\beta_2 (\cos(\Delta\theta_j) + 1)} d(\vz, \vz_{j})^2 + 1} \leq  \nonumber \\
&\max_{\vx_t \in \calP_t(\bm{\alpha})} \max_{j} \frac{\lambda_1}{\beta_1 d(\vz, \vz_{j})^2 + 1} \leq \bar{\phi} \coloneqq \frac{\lambda_1}{\beta_1 d_\textup{min}^2 + 1},
\label{eq:condition2}
\end{align}
where $d_\textup{min}$ is the closest allowable distance to obstacles, considering the robot's physical dimensions and a safety buffer. Therefore, if the risk level satisfies condition~\eqref{eq:condition2}, i.e., $\phi \leq \bar{\phi}$, then condition~\eqref{eq:local_validity} holds, indicating that the class $\calK$ functions are locally valid.


Given that the predicted risk level~$\hat{\phi}$ follows a GMM distribution due to ensemble predictions and an analytical solution for its Conditional Value at Risk~(CVaR) exists, we employ the distributionally robust CVaR as a risk metric. For notational simplicity, we slightly abuse the notation in this section by treating the risk level~$\phi$ as the output vector $Y$. For $\hat{\phi} \sim \vE_b(X)$, we define the Value at Risk~(VaR) as:
\begin{equation}
\textnormal{VaR}_{\epsilon}^{\vE_b(X)}(\hat{\phi}) \coloneqq \inf \{\nu \in \RealSpace \, | \, \textnormal{Prob}^{\vE_b(X)}(\hat{\phi} > \nu) \leq \epsilon \}.
\end{equation}
VaR represents the potential risk level with the allowable probability $\epsilon$. We then define the CVaR:
\begin{equation}
\textnormal{CVaR}_{\epsilon}^{\vE_b(X)}(\hat{\phi}) \coloneqq \inf \left\{\eta \in \RealSpace \, | \,  \eta + \frac{1}{\epsilon} \mathbb{E}_{\vE_b(X)} \left[(\hat{\phi}-\eta)^+\right] \right\},
\end{equation}
where $(\cdot)^{+} \coloneqq \max \{\cdot, 0\}$. CVaR is interpreted as the expected value over VaR, and it adheres to a group of axioms crucial for rational risk assessment~\cite{majumdar_how_2020}.

Finally, we compute distributionally robust CVaR by treating the ensemble set of distributions~$\mathfrak{E}$ as the ambiguity set~\cite{rahimian_distributionally_2022, ryu_integrating_2024}:
\begin{align}
    \sup_{\vE_b \in \mathfrak{E}} & \textnormal{CVaR}_\epsilon^{\vE_b(X)}(\hat{\phi}) \leq \bar{\phi} \Rightarrow \sup_{\vE_b \in \mathfrak{E}} \textnormal{VaR}_\epsilon^{\vE_b(X)} (\hat{\phi}) \leq \bar{\phi}  \label{eq:cvar_constraint} \\
    &\Leftrightarrow \inf_{\vE_b \in \mathfrak{E}} \textnormal{Prob}^{\vE_b(X)} (\hat{\phi} \leq \bar{\phi}) \geq 1-\epsilon.
    \label{eq:distributionally_robust}
\end{align}
By \eqref{eq:distributionally_robust}, the constraint $\hat{\phi} \leq \bar{\phi}$ is deemed as distributionally robust with the probability of $1-\epsilon$.

By evaluating these two uncertainties, \eqref{eq:epistemic_uncertainty} and \eqref{eq:cvar_constraint}, we can ensure that the prediction from the PENN model is both \textbf{\textit{confident}} and satisfies the \textbf{\textit{local validity}} condition~\eqref{eq:local_validity}.

\subsection{Online Parameter Adaptation \label{subsec:online}}

\begin{algorithm}[t]\footnotesize
\caption{Online Adaptive MPC-ICCBF}
\label{alg:online_adaptation}
\textbf{Given:} $\bm{\alpha}^{*}_0$: Initial class $\calK$ functions;  $\pi_{(\cdot)}$: MPC-ICCBF controller;\\ 
$\vF$: PENN model with trained parameters~$\bm{\theta}_{1:B}$;\\
\For{$t=1, \ldots, T_{\textup{max}}$}
{
    $\vx_t, \calE_t \leftarrow \texttt{Sensor()}$; \algovspace \\
    $\mathfrak{A}_{\textup{cand}} = \{\bm{\alpha}^{(k)}\}^{K}_{k=1} \leftarrow \texttt{SampleCandidates}(\bm{\alpha}^{*}_{t-1})$;\algovspace\\
    $\vX =\{X^{(k)} \}^{K}_{k=1} \leftarrow \texttt{CreateInputs}(\vx_t, \calE_t, \mathfrak{A}_{\textup{cand}})$; \algovspace \\
    $\mathfrak{E}, \hat{\vY} =\{\hat{Y}^{(k)} \}^{K}_{k=1} \leftarrow \vF(\vX; \bm{\theta}_{1:B})$; \algovspace\\
    $\mathfrak{A}_{\textup{valid}} = \{\bm{\alpha} : D(X; \mathfrak{E}) < D_{\textup{thr}}$ \algovspace\\
    $\qquad \qquad \qquad \qquad \qquad \wedge \, \sup_{\vE_b \in \mathfrak{E}} \textnormal{CVaR}_\epsilon^{\vE_b(X)}(\hat{\phi}) \leq \bar{\phi} \}$;\algovspace\\
    $\bm{\alpha}^{*}_t \leftarrow \argmin_{\bm{\alpha} \in \mathfrak{A}_{\textup{valid}}} \hat{\delta}$;\algovspace\\
    $\vu_t \leftarrow \pi_{\bm{\alpha}^{*}_t}(\vx_t)$;\algovspace\\
    $\texttt{ApplyControlInput}(\vu_t)$; \algovspace
}
\end{algorithm}

Integrating the concepts presented in Sec.~\ref{subsec:safety-loss}-\ref{subsec:assess}, we now introduce an online adaptive MPC-ICCBF algorithm (see Alg.~\ref{alg:online_adaptation}) that dynamically adapts the class $\calK$ functions to ensure safety and optimize performance in robot navigation scenarios.

Initially, for $t=0$, we set $\bm{\alpha}^{*}_0 = \{\gamma_{0,0}^{*}, \gamma_{1,0}^{*}\}$ to the numerically minimum values required to satisfy $\vx_0 \in \calC^{*}(\bm{\alpha})$ as per Lemma~\ref{lemma:iccbf}. At each subsequent time step, the algorithm generates $K$ candidate class $\calK$ function parameters~$\mathfrak{A}_{\textup{cand}} = \{\bm{\alpha}^{(k)}\}^{K}_{k=1}$ by sampling uniformly around the previous optimal values. These candidates form a batch of inputs~$\vX =\{X^{(k)} \}^{K}_{k=1}$ for the PENN model~$\vF$. The model then predicts GMM distributions of the risk level and the deadlock time for each input. To identify locally valid class $\calK$ functions~$\mathfrak{A}_{\textup{valid}}$, the algorithm filters these predictions based on two criteria as described in Sec.~\ref{subsec:assess}.

To approximately solve Problem~\ref{prob:optimize_performance}, the algorithm selects the class $\calK$ functions $\bm{\alpha}^{*}_t$ from $\mathfrak{A}_{\textup{valid}}$ that has the minimum predicted deadlock time $\hat{\delta}$. This minimizes the chance of overly restrictive CBF constraints impeding navigation~\cite{grover_deadlock_2021}. These parameters are then updated in the MPC-ICCBF~\eqref{eq:mpc-iccbf-formulation} problem to compute the control input~$\vu_t$. By repeating this process at each time step, it continuously refines the ICCBF parameters based on the current state and the predictions from the PENN model, thereby optimizing performance while maintaining safety.

\begin{remark} 
Assuming the discretized time step~$\Delta t$ in the MPC is sufficiently small such that $\vx_{t+1} \in \calP_t(\bm{\alpha}^{*}_{t})$. By continuously updating the locally valid class $\calK$ functions~$\bm{\alpha}^{*}_{t} \in \mathfrak{A}_{\textup{valid}}$, the system remains safe by the controller $\pi_{\bm{\alpha}^{*}_{t}} \in K_\textup{iccbf}(\vx_t; \bm{\alpha}^{*}_{t})$ rendering the set $\calP_t(\bm{\alpha}^{*}_{t}) \subset \calC^{*}(\bm{\alpha}^{*}_{t})$ forward invariant.
\end{remark} 

\section{RESULTS \label{sec:experiments}}
\subsection{Experimental Setup}

\begin{figure}[tbp]
\centering
\includegraphics[width=0.99\linewidth]{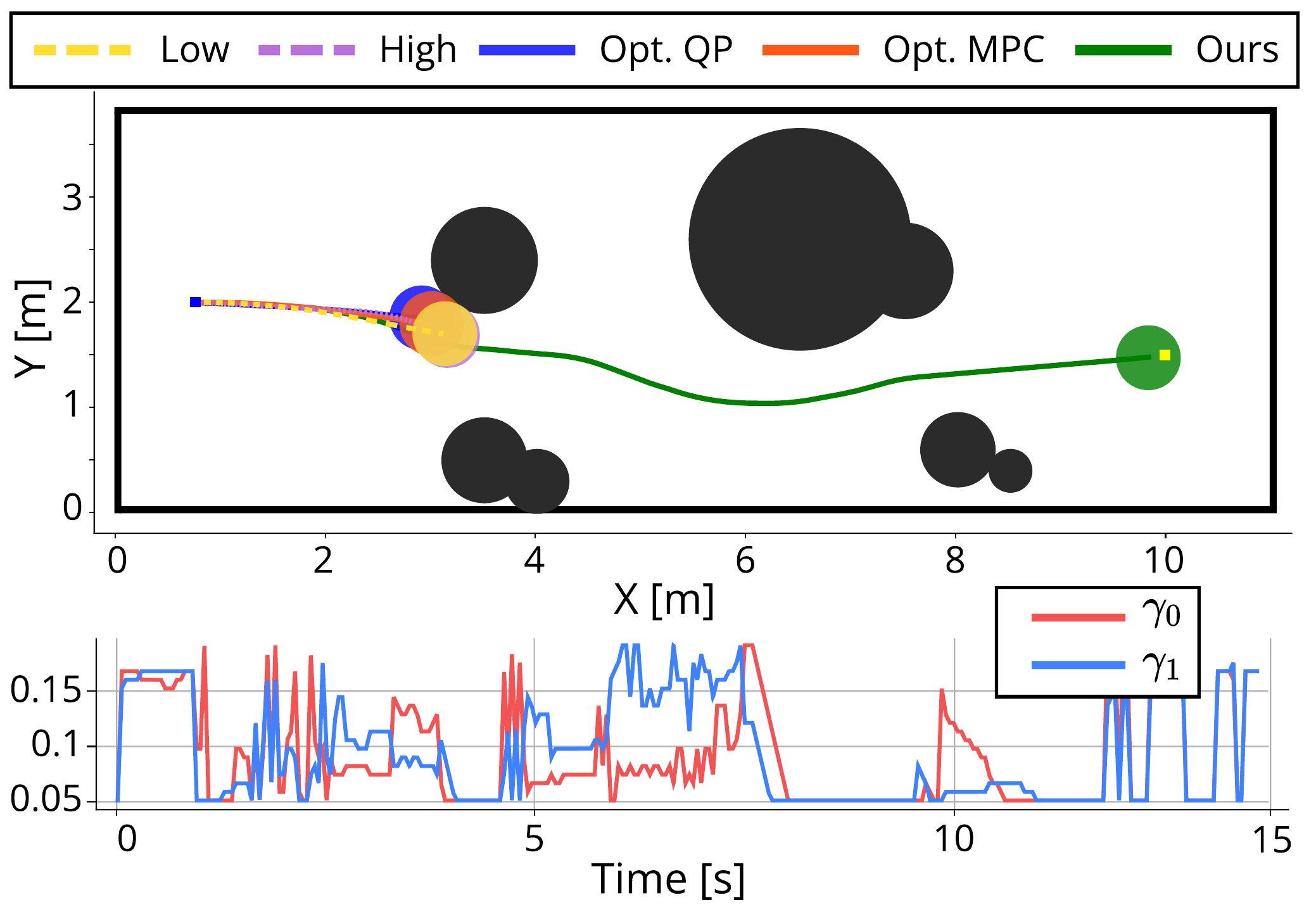}
\caption{
Visualization of robot trajectories generated by five different approaches and class $\calK$ function parameters refined over time by our online adaptive MPC-ICCBF method. The blue and yellow squares represent the start and goal location. The black circles represent the obstacles.}
\label{fig:trajectories_1}
\end{figure}

\begin{figure}[tbp]
\centering
\includegraphics[width=0.99\linewidth]{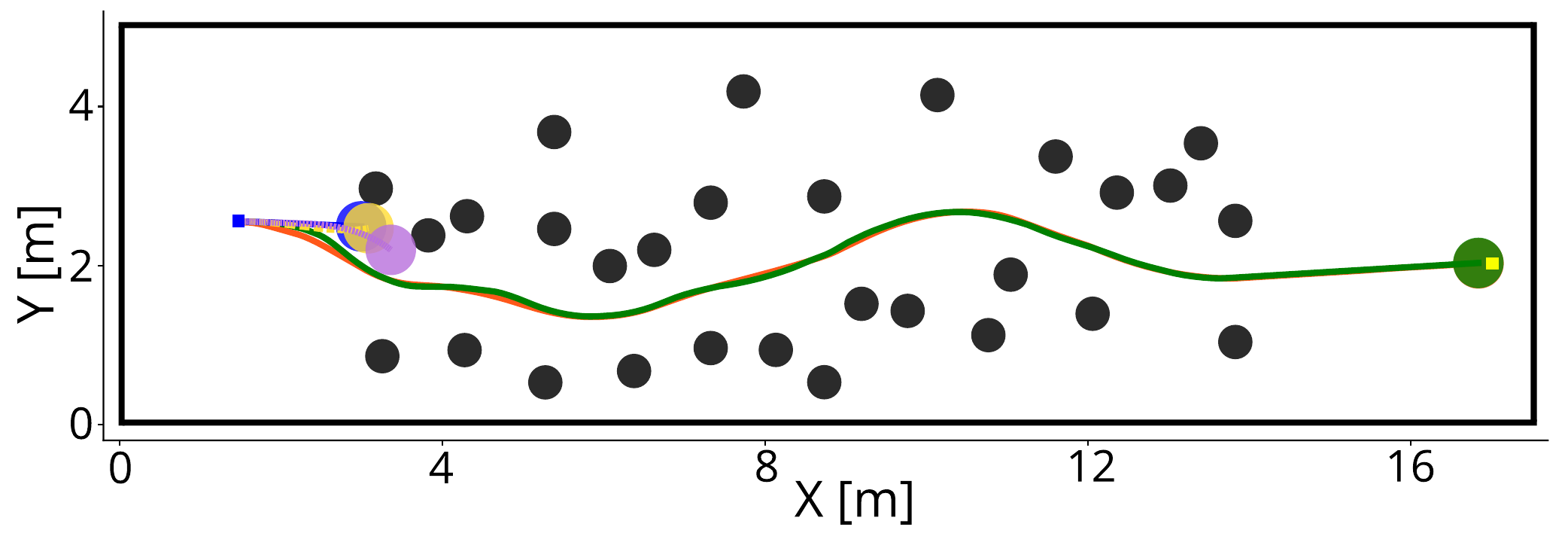}
\caption{Visualization of robot trajectories in a more complex environment.}
\label{fig:trajectories_2}
\vspace{-10pt}
\end{figure}

We consider a system has constrained inputs~\eqref{eq:mpc-iccbf-state-input-constraint} as $a \in [-0.5, 0.5]$ $\text{m/s}^2$ and $\omega \in [-0.5, 0.5]$ rad/s. The discretization time step and the MPC sampling time are set to $\Delta t = 0.05$~s. The quadratic costs in~\eqref{eq:mpc-iccbf-cost} are tuned such that the controller converges to the goal state without severe overshooting when there are no obstacles.

For training data, we generate trajectories with $N=1$ obstacle in this paper. The robot and obstacle radii are set to $l_{\textup{robot}} = 0.3$~m and $l_{\textup{obs}} = 0.2$~m, respectively. The probability~$\epsilon$ in CVaR constraint~\eqref{eq:cvar_constraint} is set to $0.05$. In total, we generate $M=16,808$ training samples offline. To simulate observation and model uncertainties, we add 3\% of i.i.d. Gaussian noise to $d_0, \Delta \theta_0, v_0$ in the input data~$X$. The PENN model~$\vF$ comprises $B=3$ ensemble models, each has 5 MLP layers with ReLU activation function. The rest of the implementation details follow \cite{kim_bridging_2023}. More implementation details can be found in our public code repository.
\subsection{Online Adaptive MPC-ICCBF}

We compare the performance of our method with four different approaches: MPC-ICCBF using 1) ``Low" ($\gamma_{0}=\gamma_{1}=0.01$) and 2) ``High" ($\gamma_{0}=\gamma_{1}=0.2$) \textit{fixed} class $\calK$ function parameters, 3) Optimal-decay CBF-QP~\cite{zeng_safetycritical_2021_decay} that can optimize class $\calK$ functions \textit{online} (abbreviated as ``Opt. QP"), and 4) ``Optimal-decay MPC-CBF"~\cite{zeng_enhancing_2021} where it incorporates the same optimization technique as \cite{zeng_safetycritical_2021_decay} within the MPC structure (abbreviated as ``Opt. MPC"). For 3) and 4), we optimize both $\gamma_{0}$ and $\gamma_{1}$ and penalize them in the cost function. For our method, we empirically set the adaptation range as $\gamma_{0}, \gamma_{1} \in (0.01, 0.2)$. As demonstrated in \cite{agrawal_safe_2021}, when the relative degree of the constraint function $h$ with respect to the control input matches the order to which the ICCBF is constructed, the ICCBF trivially reduces to an HOCBF in continuous time. Therefore, we adopt a discrete-time HOCBF implementation in our experiments.

\begin{table}[t]
\centering
\caption{Performance comparison of five approaches across two environments. Metrics include collision rate, goal-reaching rate, and average reach time. Environment \#1 corresponds to Fig.~\ref{fig:trajectories_1}, and environment \#2 to Fig.~\ref{fig:trajectories_2}. }
\begin{tabular}{l|rr|rr|rr}
\toprule
     & \multicolumn{2}{c|}{\textbf{Collision Rate}} & \multicolumn{2}{c|}{\textbf{Reach Rate}} & \multicolumn{2}{c}{\textbf{Time [s]}} \\
     & \#1 & \#2 & \#1 & \#2 & \#1 & \#2 \\
\midrule
Low      & 10.2\% & 1.9\% & 89.8\% & 98.1\%  & 54.4 & 129.0  \\
High     & 100\% & 100\% & 0\% & 0\%  & N/A  & N/A \\
Opt. QP~\cite{zeng_safetycritical_2021_decay} & 0\%  & 76.9\% & 0\% & 0\% & N/A & N/A \\
Opt. MPC~\cite{zeng_enhancing_2021} & 13.0\% & 1.9\% & 87.0\% & 98.1\% & 51.8 & 121.4 \\
\rowcolor{gray!50} Ours & 0\% & 0\% & 100\% & 100\% & 15.0 & 40.7 \\
\bottomrule
\end{tabular}
\vspace{-6pt}
\label{tab:comparison}
\end{table}

We test these approaches in two different environments. Fig.~\ref{fig:trajectories_1} and Fig.~\ref{fig:trajectories_2} illustrate the trajectories generated by each method. To quantitatively assess the effectiveness of our method, we further conduct $108$ navigation trials with uniformly varied initial states~$\vx_0$ in both environments. We evaluate three metrics: collision rate with obstacles, goal-reaching rate, and average reach time when the robot successfully reaches the goal (see Table~\ref{tab:comparison}). 

The conservative approach, using ``Low" ICCBF parameters, is significantly affected by the initial nearby obstacle due to the ICCBF constraints. In some cases, the high initial velocities~$v_{0} \in \vx_{0}$ result in states that fall outside the inner safe set defined by fixed class $\calK$ functions, leading to collisions. The aggressive approach with ``High" ICCBF parameters becomes infeasible in early iterations and eventually collides with obstacles. While ``Opt. QP" shows fewer collisions, it is prone to deadlock. ``Opt. MPC" avoids infeasibility but still results in collisions, as it only guarantees the point-wise feasibility of the solution rather than persistent safety. In contrast, our proposed method actively refines the class $\calK$ function parameters online based on the PENN model's predictions. It successfully reaches the goal without collisions or deadlocks, achieving the highest goal-reaching rate in both scenarios and demonstrating its ability to optimize performance while ensuring safety.

We also conducted hardware experiments comparing the five methods. These experiments, along with more detailed analysis, can be found in our accompanying video.

\section{CONCLUSION \label{sec:conclusion}}

In this paper, we present an online class $\calK$ function parameters adaptation approach in discrete-time ICCBFs. To this end, we first introduce the concept of locally valid class $\calK$ functions, which ensure forward invariance for a smaller subset of the inner safe set. By leveraging a PENN model with a novel two-step verification process that accounts for both epistemic and aleatoric uncertainties, our method can dynamically refine the ICCBF parameters to ensure safety and optimize MPC performance. Experimental results demonstrate that it outperforms the compared baselines in collision avoidance and goal-reaching rates across various scenarios. Future work may focus on enhancing multi-obstacle handling through graph neural network feature encoding, which could improve scalability to varying numbers of obstacles. Also, extending our method to higher-dimensional systems, such as quadrotors, is another area of interest.

\addtolength{\textheight}{0 cm}   





\bibliographystyle{IEEEtran}
\typeout{}
\bibliography{references.bib}

\end{document}